\documentclass[conference]{IEEEtran}

\IEEEoverridecommandlockouts
\usepackage{cite}
\usepackage{amsmath,amssymb,amsfonts}
\usepackage{algorithmic}
\usepackage{graphicx}
\usepackage{textcomp}
\usepackage{xcolor}
\usepackage{soul}

\usepackage{dcolumn}
\usepackage{float}
\usepackage{caption}
\usepackage{tabularx}
\usepackage{subcaption}

\usepackage{stfloats}

\def\BibTeX{{\rm B\kern-.05em{\sc i\kern-.025em b}\kern-.08em
    T\kern-.1667em\lower.7ex\hbox{E}\kern-.125emX}}

\usepackage[english]{babel}

\usepackage{amsmath}
\usepackage{amssymb}
\usepackage{amsfonts}
\usepackage{mathtools}
\usepackage[binary-units=true]{siunitx}
\usepackage{caption}
\usepackage{subcaption}
\usepackage{graphicx}
\usepackage{xcolor}

\usepackage{booktabs}
\usepackage{tabularx}

\usepackage{tikz}
\usetikzlibrary{external,automata,trees,shadows,matrix,fit,decorations,positioning,shapes,plotmarks,patterns,arrows,calc,quotes,tikzmark,arrows.meta}

\makeatletter
\renewcommand{\vec}[1]{%
	\ifcat\relax\noexpand#1%
	\ensuremath{\boldsymbol{\lowercase{#1}}}%
	\else
	\ensuremath{\mathbf{\lowercase{#1}}}%
	\fi
}
\makeatother

\makeatother

\makeatletter

\newcommand{\argmin}[1]{\ensuremath{\underset{#1}{\text{argmin }}}}

\newcommand{\argmax}[1]{\ensuremath{\underset{#1}{\text{argmax }}}}

\usepackage{pifont}

\begin{document}

\title{Conditional Gumbel-Softmax for constrained feature selection with application to node selection in wireless sensor networks\\}

\author{Thomas Strypsteen and Alexander Bertrand \IEEEmembership{}

\thanks{This project has received funding from the European Research Council (ERC) under the European Union’s Horizon 2020 research and innovation programme (grant agreement No 802895 and grant agreement 101138304). Views and opinions expressed are however those of the author(s) only and do not necessarily reflect those of the European Union or ERC. Neither the European Union nor the granting authority can be held responsible for them. The authors also acknowledge the financial support of the Flemish Government under the “Onderzoeksprogramma Artifici\"ele Intelligentie (AI) Vlaanderen” programme.
\newline
T. Strypsteen and A. Bertrand are with KU Leuven, Department of Electrical Engineering (ESAT), STADIUS Center for Dynamical Systems, Signal Processing and Data Analytics and with Leuven.AI - KU Leuven institute for AI, Kasteelpark Arenberg 10, B-3001 Leuven, Belgium (e-mail: thomas.strypsteen@kuleuven.be, alexander.bertrand@kuleuven.be).}
}

\maketitle

\begin{abstract}

In this paper, we introduce Conditional Gumbel-Softmax as a method to perform end-to-end learning of the optimal feature subset for a given task and deep neural network (DNN) model, while adhering to certain pairwise constraints between the features. We do this by conditioning the selection of each feature in the subset on another feature. We demonstrate how this approach can be used to select the task-optimal nodes composing a wireless sensor network (WSN) while ensuring that none of the nodes that require communication between one another have too large of a distance between them, limiting the required power spent on this communication. We validate this approach on an emulated Wireless Electroencephalography (EEG) Sensor Network (WESN) solving a motor execution task. We analyze how the performance of the WESN varies as the constraints are made more stringent and how well the Conditional Gumbel-Softmax performs in comparison with a heuristic, greedy selection method.
While the application focus of this paper is on wearable brain-computer interfaces, the proposed methodology is generic and can readily be applied to node deployment in wireless sensor networks and constrained feature selection in other applications as well.
\end{abstract}


\section{Introduction}
\label{section: Section1}

The last few years have seen a remarkable rise in the usage of wearable, physiological sensors measuring signals ranging from temperature, to heart rate, to glucose levels in the blood. There is however, even more potential yet to be explored by combining multiple of these modalities measured by different devices on different parts of the body in the form of a body-sensor network (BSN). A BSN leverages advancements in microprocessor miniaturization and battery efficiency to organize multiple sensors in a wireless network of nodes communicating their data to one another to allow for joint analysis of this data. One specific application for BSN's can, for instance, be found in the Wireless Electroencephalography (EEG) Sensor Network (WESN) \cite{bertrand2015distributed, narayanan2019analysis}. In this case, the sensor network is composed of multiple EEG sensor nodes, each measuring neuronal activity of the brain on different areas of the scalp. The clinical relevance of EEG has been demonstrated in multiple fields ranging from epileptic seizure detection \cite{ansari2019neonatal} to brain-computer interfaces (BCIs) \cite{lawhern2018eegnet}. Nevertheless, its application outside of hospital or laboratory settings has been severely limited by the traditional way EEG had to be measured: a bulky EEG cap with a large amount of wires connected to an acquisition device. The much more user-friendly WESN setup thus provides a promising avenue towards continuous neuromonitoring in daily life.
\newline



A major consequence of this shift towards wearable EEG solutions such as WESNs is that, in practice, we will only be able to employ a handful of channels compared to the full, dense channel layout of the cap. As such, the first step in the design of the WESN will be selecting the subset of nodes that retain the highest possible task accuracy. However, a second, crucial consideration in the design of these sensor networks is ensuring an acceptable battery lifetime. The energy bottleneck in BSNs will typically be found in the wireless transmission of the data between the sensors \cite{bertrand2015distributed,hanson2009body}. The amount of energy this will require depends not only on the amount of data to be transmitted, but is also heavily influenced by the distance between the communicating nodes, in particular when there is a 'watery' medium in between such as a human body or brain \cite{benaissa2016characterization,liao2021path} that causes a high path loss. As such, the node selection should not only be optimized for task accuracy, but also ensure the distance between nodes that will have to communicate with each other in the WESN is limited. This driver application is only one notable example where such distance constraints occur, but they pose a challenge to node deployment in wireless sensor networks in general. Larger-scale networks traditionally deal with an inherent trade-off between coverage and node distance, posing challenges for the node energy consumption, link reliability and data transfer time \cite{celis2020design, mao2019analysis, farsi2019deployment}.
\newline

To enable task-optimal node selection while adhering to these distance constraints, we introduce our novel \textit{Conditional Gumbel-Softmax} methodology. While traditional feature selection methods based on Gumbel-Softmax assume independent distributions of the features to be selected, we demonstrate that by leveraging conditional distributions, we can ensure that only node configurations are sampled that are permitted under a set of given constraints. While our evaluation use case is focused on WESNs, the methodology is generic and can readily be applied to other kinds of wireless sensor networks as well. 
\newline

The paper is organized as follows. Section \ref{section: Section2} discusses traditional and Gumbel-Softmax based feature selection approaches. Section \ref{section: Section3} introduces our conditional Gumbel-Softmax framework and how it can be applied to solve the constrained node selection problem. In Section \ref{section: Section4} we evaluate the performance of this method on an emulated WESN solving a motor execution task and discuss these in detail in Section \ref{section: Section5}.
\newline

\section{Related work}
\label{section: Section2}

\subsection{Feature selection}
\label{section: Section2a}

The goal of feature selection is to find an optimal subset of an available set of features that maximizes the performance of a classification or regression model on a given task. Filter-based approaches rank the available features by a criterion like mutual information (MI) with the target labels and select the $K$ highest scoring features \cite{lan2006salient}. Wrapper-based approaches train the model on multiple candidate feature subsets and finally select the one that performs best on a validation set. To avoid the combinatorial explosion of possibilities, such wrapper methods use heuristic methods such as greedy backwards selection to efficiently explore the space of possible feature subsets \cite{narayanan2019analysis}. Finally, embedded approaches learn the subset and the task model in an end-to-end way by, for instance, performing $L_1$ regularization on the input weights \cite{scardapane2017group}. More recent forms of embedded methods aim to perform feature selection by directly learning the discrete parameters of the selection operation through reinforcement learning \cite{yoon2018invase} or gradient estimators based on continuous relaxations \cite{paulus2020gradient}. 
\newline

While all of the above methods have shown their value for unconstrained feature selection, filter and wrapper approaches will generally have a hard time incorporating constraints in their selections. Since they typically add features to the set one by one, the constraints might cause the final selection to be very dependent on the initial feature, at which point no interactions between features could have been taken into account. Thus, when constraints are involved, the embedded approaches, that learn the entire optimal feature set jointly instead of successively, become especially suitable. Of these embedded approaches, we will focus our efforts on continuous relaxation through the Gumbel-Softmax estimator \cite{jang2016categorical,maddison2016concrete}, which allows for continuous approximations of one-hot binary vectors. We will first proceed to delve deeper into how Gumbel-Softmax operates and has been leveraged for end-to-end learnable, unconstrained feature selection \cite{abid2019concrete,singh2020fsnet,strypsteen2021end}, before showing how it can be extended to incorporate the constraints of the sensor network.

\subsection{Gumbel-Softmax based feature selection}
\label{section: Section2b}

Take a discrete random variable, drawn from a categorical distribution with $K$ classes and class probabilities $\pi_1,...\pi_N$, represented as a one-hot vector $\mathbf{\bar{z}} \in \{0,1\}^{N}$, with the index of the one indicating the class $\mathbf{\bar{z}}$ belongs to. Discrete samples from this distribution can then be drawn with the Gumbel-Max trick \cite{jang2016categorical}:
\begin{equation}
    \mathbf{\bar{z}}=\text{one\_hot}(\argmax{n} (\log \pi_n + g_n))
\end{equation}
with $g_n$ independent and identically distributed (i.i.d.) samples from the Gumbel distribution \cite{gumbel1948statistical} and $\text{one\_hot}(i)$ the operator that generates a one-hot $N\times1$ vector where the one is placed at position $i$. The Gumbel-Softmax is then a continuous, differentiable relaxation of this discrete sampling procedure, approximating the discrete one-hot vectors $\mathbf{\bar{z}}$ with continuous vectors $\mathbf{z}$ whose elements sum to one by replacing the argmax with a softmax. For the n-th element $z_n$, this results in:
\begin{equation}
\label{eq: samplingA}
    z_n = \frac{\exp((\log \ \pi_{n} + g_{n})/\tau)}{ \sum_{j=1}^{N} \exp((\log \ \pi_{j} + g_{j})/\tau)}    
\end{equation}
with $\tau$ the temperature of this continuous relaxation. This continuous relaxation of the original discrete distribution is known as the \textit{concrete distribution} \cite{maddison2016concrete}. The temperature controls the smoothness of this distribution: lower temperatures cause the samples to be increasingly one-hot and close to the discrete samples, but also causes the variance of the gradients to be higher. A typical way to perform this trade-off is by starting training at a high temperature and annealing it to small values. In the limit of $\tau \to 0$, the concrete and discrete distribution are equal:
\begin{equation}
\label{eq: limit}
    \lim_{\tau \to 0} P(z_n=1) = \frac{\pi_n}{\sum_{j=1}^{N} \pi_j}.
\end{equation}

\begin{figure}
    \centering
    \includegraphics[trim={0cm 0 0cm 0},width = 0.45\textwidth]{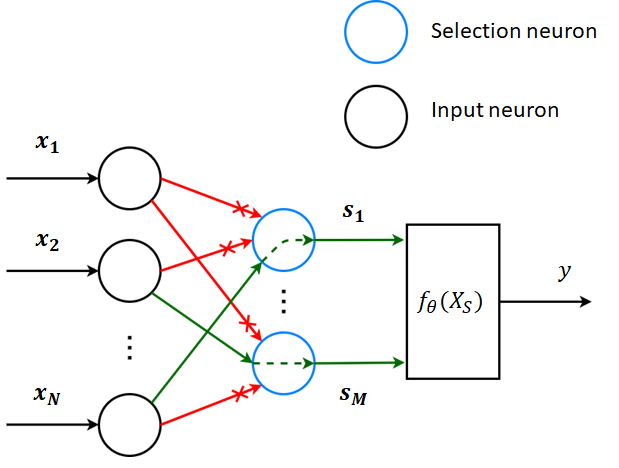}
    \caption{Schematic representation of embedded feature selection with Gumbel-Softmax. $\vec{x_n}$ indicates the feature vector derived from channel $n$. During training, the output of each selection neuron $m$ is given by $\vec{s}_m={\vec{z}^{(m)}}^{\intercal}X$, with $\vec{z}^{(m)}$ sampled from a distribution $ p_{\vec{\pi}^{(m)}}(\vec{z}^{(m)})$. The parameters $\vec{\pi}^{(m)}$ of each distribution are jointly learned with the network weights $\vec{\theta}$. During inference, each neuron only passes the input feature which had the highest probability associated with it at the end of the learning phase (illustrated by the green arrows). In traditional Gumbel-Softmax, each $\vec{z}^{(m)}$ is sampled from an independent concrete distribution as described in Eq. (\ref{eq: lossM}). In our conditional Gumbel-Softmax, $\vec{z}^{(m)}$ is sampled from a conditioned concrete distribution as described in Eq. (\ref{eq: ancestral}). Figure adapted from \cite{strypsteen2021end}. }
    \label{fig: selectionlayerscheme}
\end{figure}

The continuous relaxation allows us to approximate sampling from discrete distributions in a continuous, differentiable way, thus allowing for discrete parameters to be learned through standard backpropagation. For instance, before discussing the general case of selecting $M$ out of $N$ features, say we want to determine the single optimal feature for a given task and neural network model $f_{\vec{\theta}}$. We have a dataset consisting of samples $X \in {\rm I\!R}^{N \times L}$, with $N$ the amount of features and $L$ the dimensionality of those features, and class labels $y$. We then aim to solve the following optimization problem:
\begin{equation}
\label{eq: lossdiscrete}
    \mathbf{\bar{z}}^*, \theta^* = \argmin{\mathbf{\bar{z}},\theta} 
    \mathcal{L}(f_{\theta}(\mathbf{\bar{z}}^\top X,y))
\end{equation}
with $\mathbf{\bar{z}}$ a one-hot vector of dimension $N \times 1$ and $\mathcal{L}(p,y)$ any loss function between the predicted label $p$ and the ground truth $y$, for instance the cross-entropy. To allow this optimization problem to be solved with backpropagation, we can replace the discrete $\mathbf{\bar{z}}$ with the continuous relaxation $\mathbf{z}$, parameterized by $\vec{\pi}$:
\begin{equation}
\label{eq: loss}
    \vec{\pi}^*, \vec{\theta}^* = \argmin{\mathbf{\vec{\pi}},\vec{\theta}} 
    \mathbb{E}_{\mathbf{z} \sim p_{\vec{\pi}}(\mathbf{z})}\mathcal{L}(f_{\vec{\theta}}(\mathbf{z}^\top X,y))
\end{equation}
with $p_{\vec{\pi}}(\mathbf{z})$ being sampled from as defined in Eq. (\ref{eq: samplingA}). In this way, the parameters $\vec{\pi}$ of the 'selection layer' are jointly learned with the classifier parameters $\vec{\theta}$ in an end-to-end way. During training, the model will become more and more confident in which feature to select and the distribution $p_{\vec{\pi}}(\mathbf{z})$ will become more and more centered around a single value. At inference time, the stochasticity in the network is removed and the most likely feature is always selected:
\begin{equation}
\label{eq: argmax}
    z_{n}=
    \begin{cases}
      1, & \text{if}\ n = \argmax{j} \pi_{j} \\
      0, & \text{otherwise}
    \end{cases}
\end{equation}
The same principle can also be used to select $M$ out of $N$ features simultaneously. One way to do this would be to once again learn a single one-hot vector, which encodes each possible subset of size $M$. This however, results in a distribution of dimension $N^M$ and an equal amount of parameters, which quickly becomes infeasible. Typically, this has been solved by defining a separate concrete distribution $p_{\vec{\pi}^{(m)}}(\mathbf{z}^{(m)})$ for each feature $m$ to be selected, requiring only $NM$ parameters \cite{strypsteen2021end,abid2019concrete,singh2020fsnet}. The optimization problem then becomes:
\begin{equation}
\begin{split}
\label{eq: lossM}
    \Pi^*, \vec{\theta}^* = & \argmin{\Pi,\vec{\theta}} 
    \mathbb{E}_{Z \sim p_{\Pi}(Z)} \mathcal{L}(f_{\theta}(Z^\top X,y)) \\
    z^{(m)}_{n} = & \frac{\exp((\log \ \pi_{n}^{(m)} + g_{n}^{(m)})/\tau)}{ \sum_{j=1}^{N} \exp((\log \ \pi_{j}^{(m)} + g_{j}^{(m)})/\tau)}     
\end{split}
\end{equation}
with $Z = [\mathbf{z}^{(1)},...,\mathbf{z}^{(M)}]$ and $\Pi = [\vec{\pi}^{(1)},...,\vec{\pi}^{(M)}]$. At inference time then, the selection is similar to Eq. (\ref{eq: argmax})
\begin{equation}
\label{eq: argmaxM}
    z_{n}^{(m)}=
    \begin{cases}
      1, & \text{if}\ n = \argmax{j} \pi_{j}^{(m)} \\
      0, & \text{otherwise}
    \end{cases}
\end{equation}
This process is illustrated in Fig. \ref{fig: selectionlayerscheme}. In the next section, we will discuss how our proposed Conditional Gumbel-Softmax method extends on the principles discussed in this section to allow for end-to-end learning of the optimal feature subset while adhering to pairwise constraints.

\section{Proposed Method}
\label{section: Section3}

\subsection{Conditional Gumbel-Softmax}
\label{section: Section3a}

\begin{figure}
\centering
\includegraphics[width=0.5\textwidth]{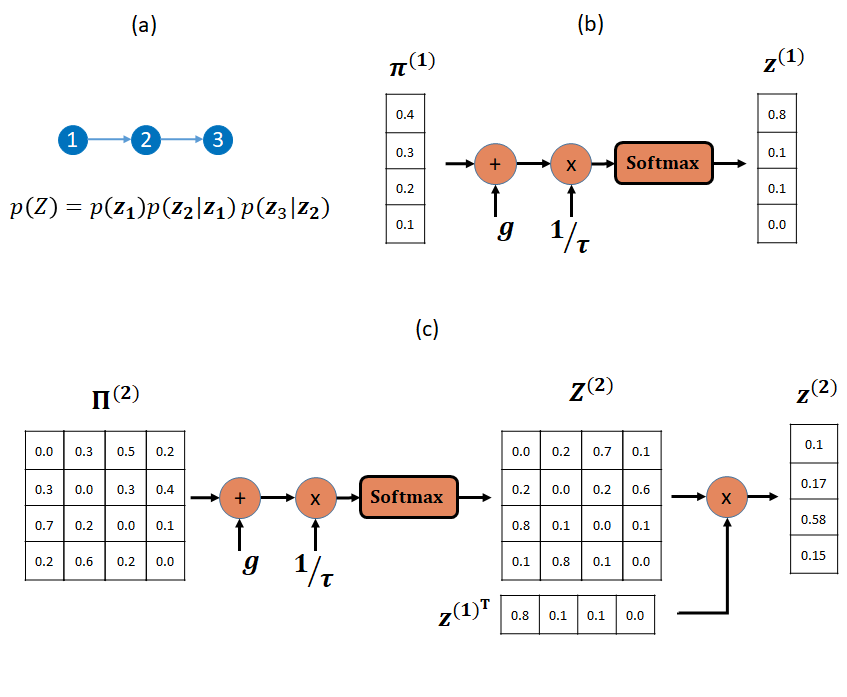}
\caption{Illustration of sampling in the conditional Gumbel-Softmax framework. In this example, the optimal 3 out of 4 features need to be selected. (a) A Bayesian network and its corresponding factorization are chosen. (b) A concrete sample $\mathbf{z}^{(1)}$ for the root vertex of the factorization is drawn through the standard Gumbel-Softmax trick of Eq. (\ref{eq: samplingA}). (c) The second vertex of the factorization is conditioned on the first one. Concrete samples $Z^{(2)}$ are drawn for every row of the conditional distribution matrix $\Pi^{(2)}$. These are then weighted by the previously drawn concrete sample of the conditioning vertex $\mathbf{z}^{(1)}$, resulting in a concrete sample $\mathbf{z}^{(2)}$. This procedure is then repeated for every entry of the desired feature subset.}
\label{fig: CGS_sampling}

\end{figure}

The usage of Eq. (\ref{eq: lossM}) implies independent sampling of the different $\mathbf{z}^{(m)}$. In other words, to make learning of the joint distribution $p(Z)$ tractable, it has been factorized in a product of the marginal distributions:
\begin{equation}
\label{eq: marginals}
p(Z) = \prod_{m} p_{\vec{\pi}^{(m)}}(\mathbf{z}^{(m)}).
\end{equation}
The core idea of our proposed \textit{Conditional Gumbel-Softmax} is to factorize this distribution in a different way, by conditioning the distributions of a feature $m$ on one different feature $m'$. To keep the amount of parameters manageable, we let each feature only be conditioned on a single other feature. Without loss of generality, assume $\mathbf{z}^{(1)}$ is the root of this factorization, i.e., the feature whose distribution is not conditioned on any other. This new factorization then takes the form:
\begin{equation}
\label{eq: factorization}
    p(Z) = p_{\vec{\pi}^{(1)}}(\mathbf{z}) \prod_{m=2}^{M} p_{\Pi^{(m)}}(\mathbf{z}^{(m)} | \mathbf{z}^{(m')})
\end{equation}
with $m'$ the feature index on which feature $m$ is conditioned. This factorization is parametrized by one vector $\vec{\pi}^{(1)} \in {\rm I\!R}^{N}_{>0}$ for the marginal probability of the root $\mathbf{z}^{(1)}$ and $M-1$ matrices $\Pi^{(m)} \in {\rm I\!R}^{N \times N}_{>0}$. This factorization can be represented by a Bayesian network, which, since each node in the network is only conditioned on a single other node at most, takes the shape of a polytree. Sampling from this distribution then happens in a manner similar to ancestral sampling. For each node $m$, a concrete sample is drawn from each row of the conditional probability matrix $\Pi^{(m)}$. These samples are then weighted based on the concrete sample drawn from the predecessor node $m'$:
\begin{equation}
\begin{split}
\label{eq: ancestral}
    Z_{kn}^{(m)} = & \frac{\exp((\log \ \Pi_{kn}^{(m)} + g_{kn})/\tau)}{ \sum_{j=1}^{N} \exp((\log \ \Pi_{kj}^{(m)} + g_{kj})/\tau)} \\
    & \text{for } k=1,...,N \\
    \mathbf{z}^{(m)} = & {\mathbf{z}^{(m')}}^\top Z^{(m)}
\end{split}
\end{equation}
This sampling scheme is illustrated in Fig. \ref{fig: CGS_sampling}. As the temperature of the Gumbel-softmax decreases, this procedure more closely resembles actual ancestral sampling, as the weighting by the predecessor sample simply becomes a selection of a row of the conditional matrix.
The advantage of this new factorization is twofold. Firstly, it allows for better modeling of the joint selection distribution $p(Z)$, while keeping the amount of parameters manageable. Secondly, as we will show in the next section, it allows for the incorporation of pairwise constraints between the different elements of the selection.

\subsection{Constrained node selection in a WESN}
\label{section: Section3b}

\begin{figure}
\centering
\includegraphics[width=0.5\textwidth]{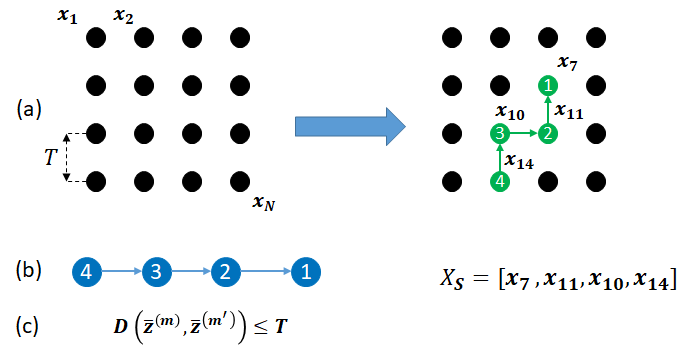}
\caption{Constrained node selection for a WSN. Given are (a) a grid of $N$ node coordinates, their pairwise distance matrix $D$ and their data $X = [\mathbf{x}_1, ..., \mathbf{x}_N]$, (b) the graph $\mathcal{G}$ representing the desired communication topology of the WSN and (c) constraints indicating a maximal allowable distance $T$ between nodes having to communicate with each other. The goal is then to find a selection of nodes $X_S = \bar{Z}^\top X$ that can be mapped to the communication topology while adhering to the constraints, while optimizing performance for the given classification/regression task.}
\label{fig: WSNproblem}

\end{figure}

Our goal is to select the task-optimal $M$ out of $N$ possible mini-EEG nodes. Each node measures a certain EEG channel and corresponds to a certain location on the scalp. In previous work \cite{strypsteen2021end}, this was done in an unconstrained way using the Gumbel-Softmax as described by Eq. (\ref{eq: lossM}). Now however, we want to take into account that these nodes will be organized in a wireless sensor network, with each sensor node relaying its measured data to another node, until all the channels have been aggregated on a single node, where inference will then happen. The topology these nodes use to communicate with each other is given and can be represented by a graph $\mathcal{G} = (V,E)$. We also assume the distances between any node pair is known and are given by the matrix $D \in {\rm I\!R}^{N \times N}_{>0}$. The constraint we want to impose on the selection is that the distance between each pair of nodes that have to communicate with each other does not exceed a certain distance, thereby limiting the energy required for the wireless communication. The task thus becomes to find a selection of $M$ nodes $\bar{Z} = [\mathbf{\bar{z}}^{(1)},...,\mathbf{\bar{z}}^{(M)}]$ with $\mathbf{\bar{z}}^{(m)}$ a $N \times 1$ one-hot vector indicating which of the $N$ nodes is selected for each vertex $v^{(m)}$ of the communication graph. If an edge $E_{ij}$ exists between two vertices $(v_i,v_j)$, the distance between the two nodes assigned to these vertices is not allowed to exceed a given threshold $T$:
\begin{equation}
\label{eq: constraint}
\forall \{(i,j) | E_{ij} = 1 \}: \mathbf{\bar{z}}^{{(i)}^\top} D \mathbf{\bar{z}}^{(j)} \leq T
\end{equation}
This situation is schematically illustrated in Fig. \ref{fig: WSNproblem}.
\newline

\begin{figure*}[htbp]
\centering
\includegraphics[width=0.9\textwidth]{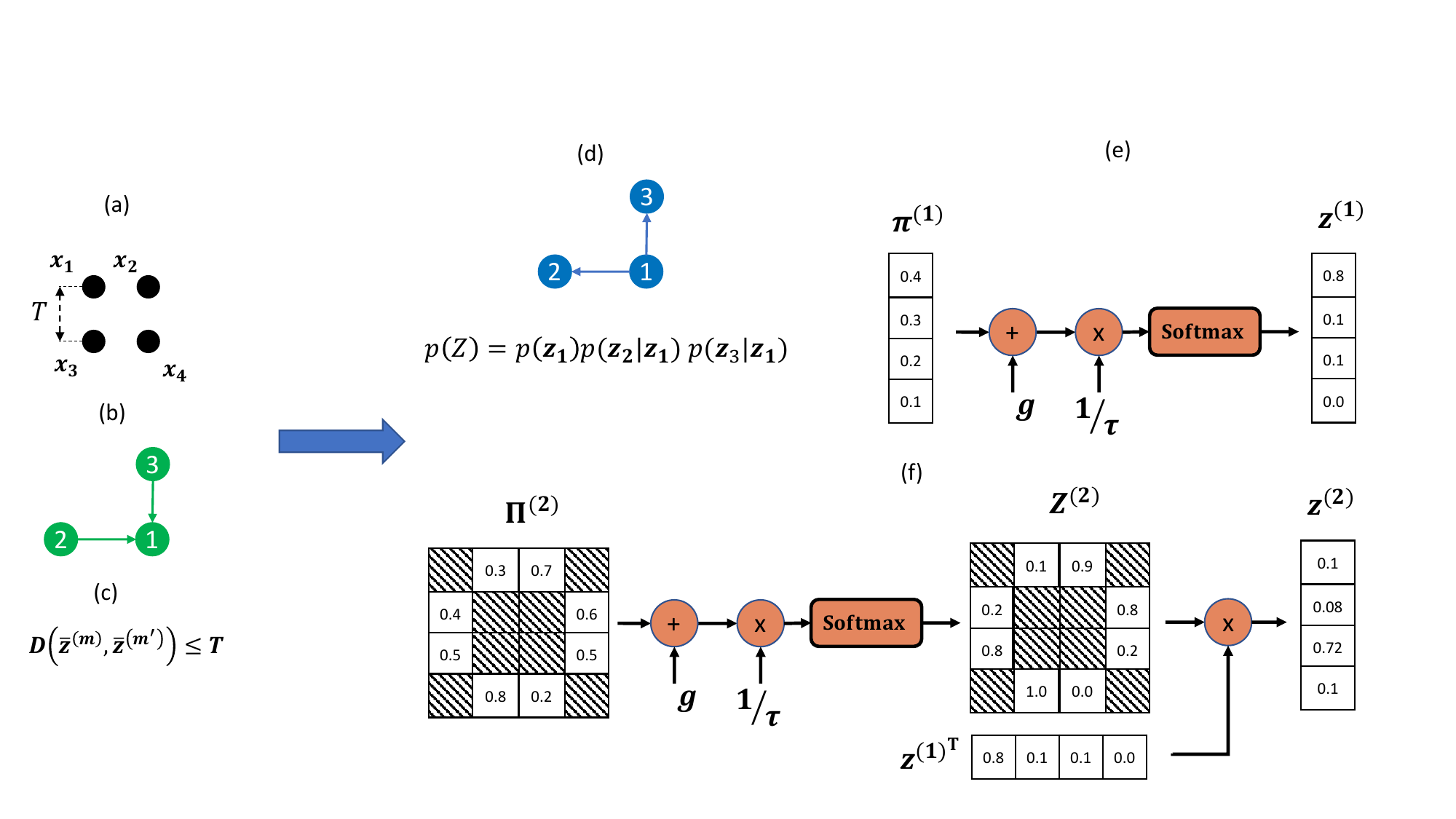}
\caption{Applying Conditional Gumbel-Softmax (d-f) to the constrained sensor node selection determined by the node layout (a), the desired communication graph (b) and the distance constraint (c). Firstly, the Bayesian network for the factorization (d) is obtained by taking the transpose of the communication graph (b). Secondly, entries of the conditional probability distributions that represent a combination of sensor nodes that are not allowed by the constraints are zeroed out (f). This guarantees that the only node configurations sampled are those that are allowed by the constraints. The diagonal elements of the conditional matrices are also zeroed to avoid selecting the same node multiple times. Sampling then occurs in the same way as described in Eq. (\ref{eq: ancestral}) and Fig. \ref{fig: CGS_sampling}. Note that while this example has a star topology as a communication graph, our method is equally applicable to line or tree topologies.}
\label{fig: CGS_WSN}

\end{figure*}

These constraints can be elegantly incorporated with the Conditional Gumbel-Softmax method by employing the given communication graph $\mathcal{G}$ as the transposed\footnote{This means that the direction of the edges are reversed.} Bayesian network for the factorization of the joint selection $Z$ in Eq. (\ref{eq: factorization}). Thus, the selection of each node $m$ is conditioned on the node $m'$ it will have to transmit its data to. To ensure that no pairs of communicating nodes that are too far away from each other are selected, we can zero the elements of the conditional distributions $\Pi^{(m)}_{ij}$ for which $D_{ij} > T$. This guarantees that the only node configurations that can ever be sampled satisfy the given constraints. This procedure is illustrated in Fig. \ref{fig: CGS_WSN}.

\section{Experimental results}
\label{section: Section4}

\subsection{Data set}
\label{section: Section4a}

We investigate the use of our Conditional Gumbel-Softmax for constrained node selection in the context of a wireless EEG sensor network solving a specific BCI task: motor execution. In such a task, the goal is to, given EEG data, decode which specific body movement gave rise to this specific EEG. To this end, we will employ the High Gamma Dataset \cite{schirrmeister2017deep}, which contains 14 subjects whose EEG was recorded while performing one of four specific movements following a visual cue: left hand, right hand, feet and rest. The training set contains 880 of such trials, of which we employ 80\% for training and the rest as a validation set for early stopping. Mean accuracies across the subjects are then reported on the held-out test set of about 160 trials per subject. The data is then preprocessed as described in \cite{schirrmeister2017deep}. We only consider the 44 channels of the motor cortex for this motor execution task. The data is resampled to 250 Hz, high-pass filtered at 4 Hz and the mean and variance of each channel is standardized to 0 and 1 respectively. Finally, the relevant time windows are extracted from the data by taking the 0.5 seconds before the visual cue and the 4 seconds after. For classifcation, we make use of the multiscale parallel filter bank convolutional neural network (MSFBCNN) proposed in \cite{wu2019parallel}. For completeness, a detailed summary of this network in table format can be found in Appendix A.
\newline

\subsection{Model performance}
\label{section: Section4b}
In this section, we analyze the performance of the proposed Conditional Gumbel-Softmax when solving the constrained channel selection problem for the motor execution dataset. To do this, we perform selection of the optimal $M=4$ out of $N=44$ channels available in the HGD dataset, while varying the maximally allowable distance $T$ between communicating nodes. The constraint is normalized such that the distance between the two furthest channels equals 1. We do this for two communication topologies: a line graph where each nodes forwards both its own measured and previously received data to the next node until the root is reached and a star graph, where all nodes send their data to the root node. Additionally, to reduce the variance of the selection layer's gradients, we repeat the sampling procedure of Eq. (\ref{eq: ancestral}) 5 times and average the resulting concrete samples for each training sample. As a baseline, we use a heuristic filter approach. We rank all the channels according to their mutual information (MI) with the class label. Then, we construct our selected feature set by iteratively adding the highest ranked channel that does not violate the constraints to this set. We also compare both these methods with the unconstrained, vanilla Gumbel-Softmax method of Eq. (\ref{eq: lossM}) as an upper limit of their performance (note that this method does not take the distance constraints into account). For a fair comparison, we also repeat the previously mentioned averaging of the concrete samples over 5 sampling rounds when using the vanilla Gumbel-Softmax. 
\newline

\begin{figure*}[!htbp]
        \centering

	\begin{minipage}{0.2\linewidth}
		\centering  
		\includegraphics[trim={0cm 0cm 0cm -1cm},clip,width=\textwidth]{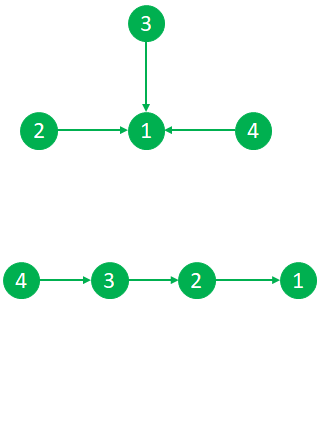}
		\caption*{(a) Star and line topology}
		\label{fig:resultsa}
	\end{minipage}%
	\begin{minipage}{0.4\linewidth}
		\includegraphics[trim={0cm 0cm 0cm 0cm},clip,width=\textwidth]{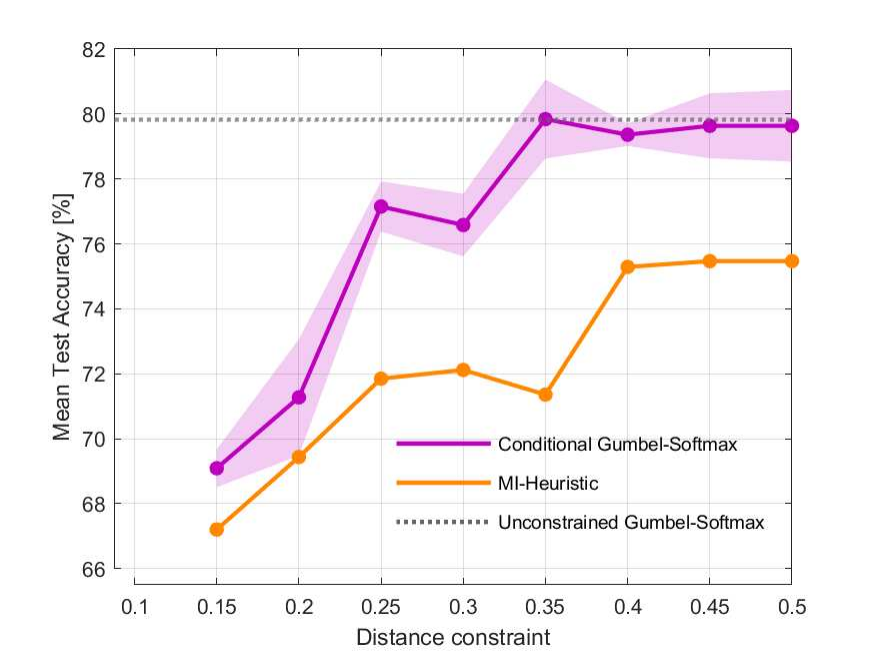}
		\caption*{(b) Star performance}
		\label{fig:resultsb}
	\end{minipage}%
	\hfill
	\begin{minipage}{0.4\linewidth}
		\includegraphics[trim={0cm 0cm 0cm 0cm},clip,width=\textwidth]{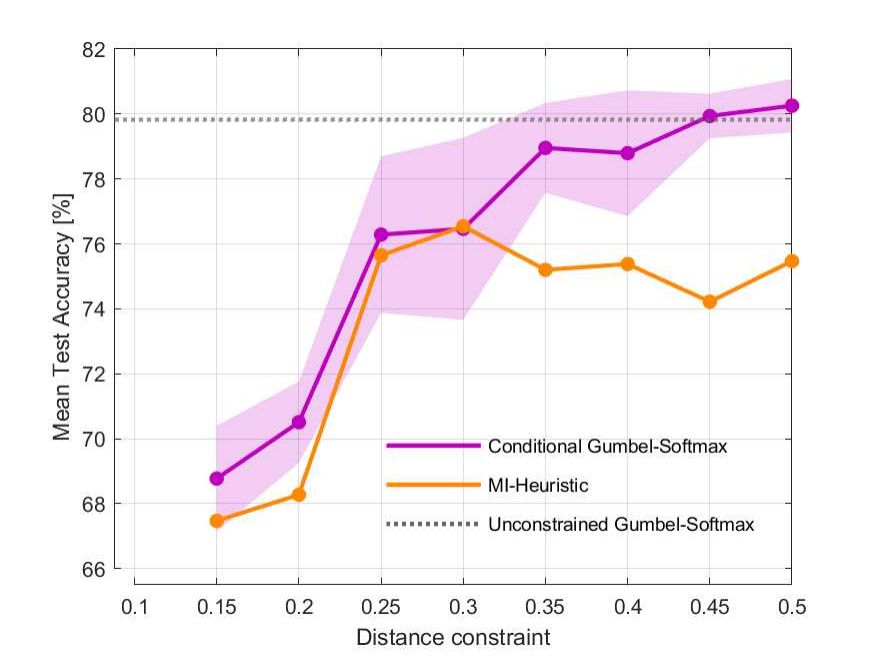}
		\caption*{(c) Line performance}
		\label{fig:resultsc}
	\end{minipage}
	\hfill
	\caption[Performance of constrained channel selection with Conditional Gumbel-Softmax for motor execution]{Performance of the Conditional Gumbel-Softmax for constrained channel selection on the motor execution task. Mean test accuracies are plotted against the distance threshold $T$ for a star and line communication topologies for $M=4$ channels. Shaded regions indicate one standard deviation of the individual runs. Dotted lines indicate accuracy of the unconstrained vanilla Gumbel-Softmax method.}
    \label{fig: accuracies}
\end{figure*}

The resulting accuracies are shown in Fig. \ref{fig: accuracies}. At very low distances, the accuracies are severely impacted and fall in the neighbourhood of what we expect when selecting only 2 channels in an unconstrained way. Once the constraints are lifted somewhat higher, the accuracy shoots up: at this point, the constraints are flexible enough such that well-performing channels from both hemispheres can be used, which is known to be beneficial for the motor execution task. As the distance constraint becomes relaxed enough, the conditional Gumbel-Softmax performs equally good as vanilla Gumbel-Softmax. In general, the conditional Gumbel-Softmax is able to find solutions that outperform the heuristic, MI-based filter approach, but its major downside is that it can suffer from high variance and can get stuck in suboptimal solutions. This can be most clearly seen for the line topology (Fig. \ref{fig: accuracies}(c)) in the interesting constraint regions around $T=0.25$ and $T=0.3$ in Fig. \ref{fig: accuracies}. In this region, configurations that have near-unconstrained accuracy exist, but are not always found. This also results in the conditional Gumbel-Softmax method sometimes getting stuck in a solution with a worse performance than the MI heuristic.

\section{Discussion}
\label{section: Section5}
We have presented Conditional Gumbel-Softmax, an extension to the Gumbel-Softmax based concrete selection layer \cite{abid2019concrete,strypsteen2021end,singh2020fsnet}, where the selection of each element of the feature subset is conditioned on another element of this subset. The main advantage of this conditioning is that it allows for the incorporation of pairwise constraints. We have demonstrated its application to constrained EEG channel selection, where the goal is to find a set of EEG channels that are both task-relevant and whose distance from each other does not exceed a given amount. These kinds of constraints fall outside the scope of traditional feature selection methods, so we compared the performance of Conditional Gumbel-Softmax with an ad hoc heuristic that selected channels with a good MI-based ranking that did not violate the constraints. The results show that while Conditional Gumbel-Softmax is fully able to find high-performing channel subsets within the constraints, it is also possible to get stuck in very suboptimal minima performing worse than the heuristic approach. To solve this issue, further research analyzing the different trajectories optimal and suboptimal solutions follow during training will be required.
\newline

It is important to note that the constrained channel selection problem statement formulated in Section \ref{section: Section3} is only a limited model of the actual problem we aim to solve: the minimization of the communication power. In the simple model presented here, the distance between two nodes is used as a surrogate for the required communication power, implicitly assuming a direct one-to-one relationship between the two. This in turn, implies that the amount of data that must be transmitted from each node is equal. It can already be seen that this is not really the case for the line topology: since each node also needs to relay the data it received from the previous node, the next to last node in the line will need to transmit a multiple of the data of the first node. In this case, this could be accounted for by imposing a different threshold $T$ for each conditional matrix $\Pi^{(m)}$. But more importantly, the actual relationship between node distance and communication power can be a lot more complex than presented here. Still, if it is possible to obtain the required communication power for each possible pair of nodes through a model or measurements, it would still be possible to use the conditional Gumbel-Softmax by replacing the distance matrix $D$ in Eq. (\ref{eq: constraint}) with this newly obtained power consumption matrix.
\newline

Finally, it would also be possible to use the conditional Gumbel-Softmax to optimize the selection for other metrics of the communication topology, such as latency. However, it is important to make a distinction between \textit{local} metrics such as communication power and \textit{global} metrics such as latency. Since each node has a local battery, each node has its own power budget, with the communication power only depending on the distance between this node and the next one in the topology. This means that selecting a certain pair of nodes to communicate with each other can violate the constraints, independent of the other nodes comprising the network. In the conditional Gumbel-Softmax framework, this means that by eliminating such pairs in the conditional matrices $\Pi^{(m)}$, we can ensure the only node selections that are sampled are those that satisfy the constraints. Latency however, is a metric that depends on the \textit{entire} network. There are thus no node pairs we can eliminate a priori, and no way we can ensure exclusive sampling from node selections that satisfy the constraints by manipulating the conditional matrices. If we want to optimize the selection for such global metrics, we will instead have to use the conditional Gumbel-Softmax in a \textit{soft} way. Instead of eliminating constraint-breaking selections during sampling, we will have to extend the supervised loss with a regularization loss that, for instance, penalizes the expected value of the latency instead.

\bibliographystyle{IEEEtran}
\bibliography{mybib.bib}

\onecolumn

\appendices

\section{MSFBCNN architecture}

\begin{table*}[!h]
\begin{tabular}{llllllll}
\hline
\textbf{Layer} & \textbf{\# Filters} & \textbf{Kernel} & \textbf{Stride} & \textbf{\# Params} & \textbf{Output} & \textbf{Activation} & \textbf{Padding} \\ \hline
Input          &                     &                 &                 &                    & (C,T)           &                     &                  \\
Reshape        &                     &                 &                 &                    & $(1,T,C)$       &                     &                  \\
Timeconv1      & $F_T$               & $(64,1)$        & $(1,1)$         & $64F_T$            & $(F_T,T,C)$     & Linear              & Same             \\
Timeconv2      & $F_T$               & $(40,1)$        & $(1,1)$         & $40F_T$            & $(F_T,T,C)$     & Linear              & Same             \\
Timeconv3      & $F_T$               & $(26,1)$        & $(1,1)$         & $26F_T$            & $(F_T,T,C)$     & Linear              & Same             \\
Timeconv4      & $F_T$               & $(16,1)$        & $(1,1)$         & $16F_T$            & $(F_T,T,C)$     & Linear              & Same             \\
Concatenate    &                     &                 &                 &                    & $(4F_T,T,C)$    &                     &                  \\
BatchNorm      &                     &                 &                 & $2F_T$             & $(4F_T,T,C)$    &                     &                  \\
Spatialconv    & $F_S$               & $(1,C)$         & $(1,1)$         & $4CF_TF_S$         & $(F_S,T,1)$     & Linear              & Valid            \\
BatchNorm      &                     &                 &                 & $2F_S$             & $(F_S,T,1)$     &                     &                  \\
Non-linear     &                     &                 &                 &                    & $(F_S,T,1)$     & Square              &                  \\
AveragePool    &                     & $(75,1)$        & $(15,1)$        &                    & $(F_S,T/15,1)$  &                     & Valid            \\
Non-linear     &                     &                 &                 &                    & $(F_S,T/15,1)$  & Log                 &                  \\
Dropout        &                     &                 &                 &                    & $(F_S,T/15,1)$  &                     &                  \\
Dense          & $N_C$               & $(T/15,1)$      & $(1,1)$         & $F_S(T/15)N_C$     & $N_C$           & Linear              & Valid            \\ \hline
\end{tabular}
\caption{Architecture of the MSFBCNN used for motor execution classification. This table is cited from \cite{wu2019parallel}.}
\end{table*}

\end{document}